\documentclass[letterpaper, 10 pt, conference]{ieeeconf}  

\IEEEoverridecommandlockouts                              
\usepackage{graphicx}
\usepackage{color,soul}
\usepackage{hyperref}
\usepackage{mwe}
\usepackage{subcaption}
\usepackage{amsmath}
\usepackage{tabularx} 
\usepackage{multirow}

\newcolumntype{L}{>{\centering\arraybackslash}m{3cm}}
\usepackage{float}
\usepackage[thinc]{esdiff}
\usepackage{graphics} 
\usepackage[table]{xcolor}
\usepackage{algorithmic}
\usepackage[ruled,norelsize]{algorithm2e} 
\usepackage{caption}
\usepackage{graphics} 
\usepackage{epsfig} 
\usepackage{mathptmx} 
\usepackage{times} 
\usepackage{amsmath} 
\usepackage{amssymb}  
\usepackage[noadjust]{cite}
\usepackage{booktabs}
\usepackage{multirow}
\usepackage{algorithmic}
\usepackage[ruled,norelsize]{algorithm2e} 
\usepackage[noadjust]{cite}

\usepackage{microtype}
\renewcommand{\baselinestretch}{0.965}

\title{\LARGE \bf
PROSPECT: Precision Robot Spectroscopy Exploration and Characterization Tool
}

\author{Nathaniel Hanson$^{1,4,\dag,*}$, Gary Lvov$^{1,\dag}$, Vedant Rautela$^{1}$, Samuel Hibbard$^{1}$, Ethan Holand$^{1,3}$,\\ Charles DiMarzio$^{2}$, and Taşkın Padır$^{1}$
\thanks{$^{\dag}$Equal Contribution; $^{*}$Correspondence: {\tt\footnotesize nhanson2@mit.edu}}%
\thanks{$^{1}$Institute for Experiential Robotics; $^{2}$Electrical and Computer Engineering Department, Northeastern University, Boston, Massachusetts, USA.}
\thanks{$^{3}$Robotics Institute, Carnegie Mellon University, Pittsburgh, Pennsylvania, USA.}
\thanks{$^{4}$Lincoln Laboratory, Massachusetts Institute of Technology, Lexington, Massachusetts, USA.}
\thanks{Ta\c{s}k{\i}n Pad{\i}r holds concurrent appointments as a Professor of Electrical and Computer Engineering at Northeastern University and as an Amazon Scholar. This paper describes work performed at Northeastern University and is not associated with Amazon.}
\thanks{Project Code, Materials List, and Assembly Instructions: \url{https://river-lab.github.io/prospect}
}
}

\begin{document}

\maketitle
\thispagestyle{empty}
\pagestyle{empty}

\begin{abstract}
Near Infrared (NIR) spectroscopy is widely used in industrial quality control and automation to test the purity and grade of items. In this research, we propose a novel sensorized end effector and acquisition strategy to capture spectral signatures from objects and register them with a 3D point cloud. Our methodology first takes a 3D scan of an object generated by a time-of-flight depth camera and decomposes the object into a series of planned viewpoints covering the surface. We generate motion plans for a robot manipulator and end-effector to visit these viewpoints while maintaining a fixed distance and surface normal. This process is enabled by the spherical motion of the end-effector and ensures maximal spectral signal quality. By continuously acquiring surface reflectance values as the end-effector scans the target object, the autonomous system develops a four-dimensional model of the target object: position in an $R^3$ coordinate frame, and a reflectance vector denoting the associated spectral signature. We demonstrate this system in building spectral-spatial object profiles of increasingly complex geometries. We show the proposed system and spectral acquisition planning produce more consistent spectral signals than na\"{\i}ve point scanning strategies. Our work represents a significant step towards high-resolution spectral-spatial sensor fusion for automated quality assessment.
\end{abstract}


\section{Introduction}
\label{sec:intro}
Quality, repeatability, and safety are key considerations for manufacturing processes. In a zero-defect manufacturing paradigm, items are selected for grading and evaluation at multiple points in the production process, with a preference for non-destructive testing when possible \cite{abasi2018dedicated}. Spectroscopy is a commonly used non-contact technique to assess manufacturing quality. By analyzing patterns of absorption or emission of electromagnetic radiation by a material, spectroscopy provides information about its composition, structure, and physical properties \cite{pasquini2018near}. Spectral information is alternatively acquired through hyperspectral imaging (HSI) -- a technique that extends RGB imaging to $n$ discrete wavelengths. Spectral sensing has many diverse applications in agriculture \cite{adao2017hyperspectral}, food processing \cite{liu2017hyperspectral}, geology \cite{van2012multi}, and remote sensing \cite{shaw2003spectral}.

Spectral data are useful for determining the \textit{material} nature of items with precision far exceeding RGB imaging, and without physical contact. Therefore, fusing spectral data with spatial information creates richer object models. For example, knowing where agricultural items are bruised or diseased allows these regions to be excised, preserving the remaining product. Additionally, knowing where specific corrosion patterns exist on a built structure helps to estimate the extent of structural decay.
\begin{figure}[t]
    \centering
    \includegraphics[width=\linewidth]{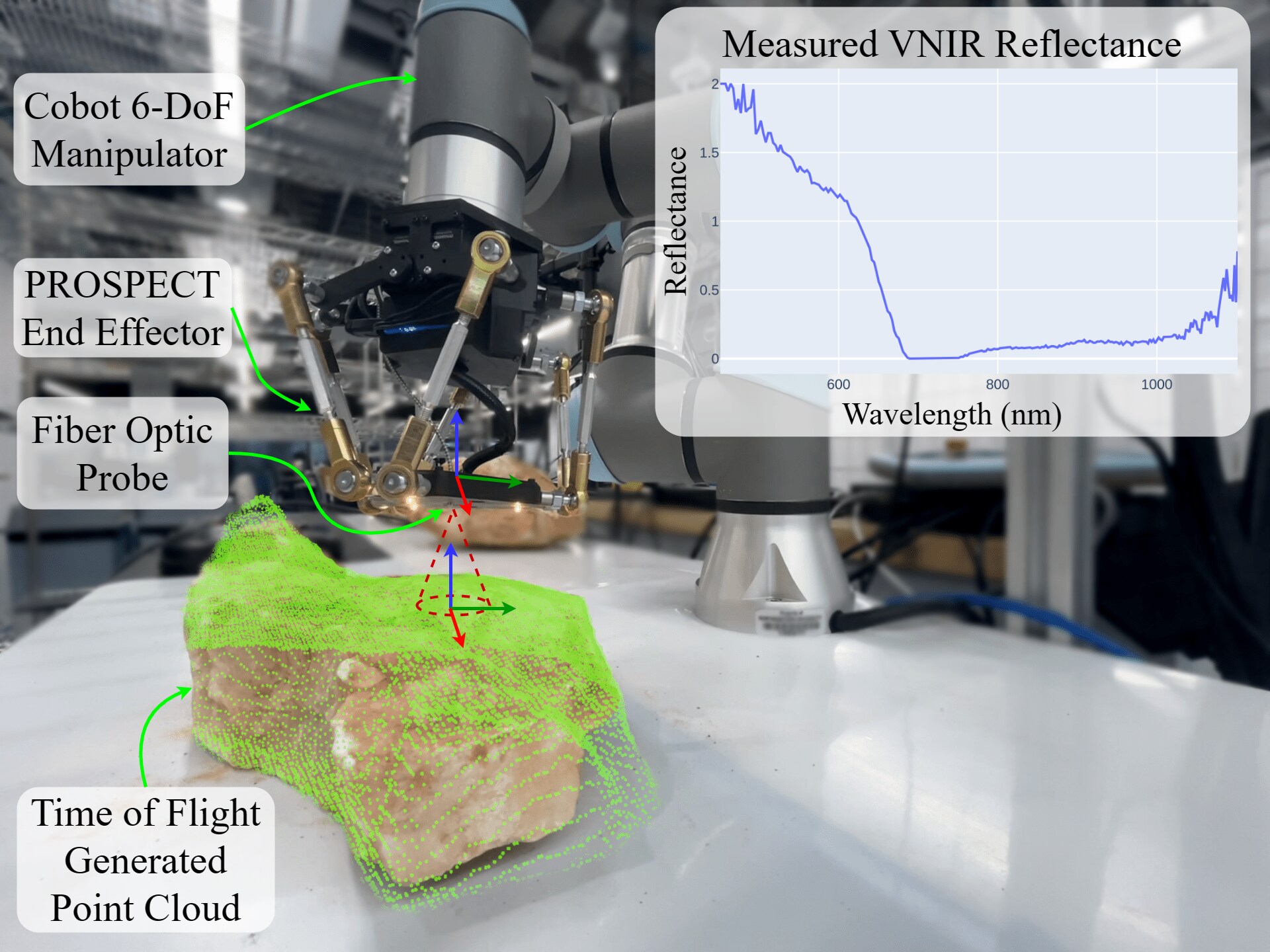}
    \caption{PROSPECT manipulator mounted to robot arm and scanning a surface with observed Visible-Near Infrared (VNIR) reflectance profile and intersecting region of points forming the signature.}
    \label{fig:prospect_teaser}
    \vspace{-1.90em}
\end{figure}

However, manual acquisition of spectral information is time intensive and requires precision alignment of a probing element to cover a surface. Similarly, HSI reduces spectral information to a 2D projection. Most HSI systems use a pushbroom acquisition, in which the camera or object must be linearly translated to acquire a full image. HSI systems also require intense active illumination for indoor operation. In contrast, point spectroscopy acquires a 1D spectral profile from the target object, but it can offer an order of magnitude increase in spectral resolution and cost-effectiveness at these points. 

Acquiring multiple spectral measurements from over a surface requires a probe to be precisely translated as signal formation is highly dependent on the distance of the sensed object from the measurement device. Thus, the core question of our work is: \textit{Can point spectroscopy coupled with a dexterous end-effector autonomously generate high-resolution spectral-spatial profiles of objects for inspection?}

In this research, we present PROSPECT, a novel, cost-effective, sensorized end-effector for precision illumination and alignment of fiber-optic coupled spectroscopy. This research focuses on the design and control of the end-effector, as well as acquisition modeling of spectral signals to construct spectral point clouds. 

The contribution of this work are as follows:
\begin{itemize}
    \item Design, kinematics modeling, and validation of an end-effector for precise point spectral measurements.
    \item An algorithm for spectral viewpoint selection to maximize signal consistency and coverage over a 3D surface. 
    \item Modeling of a spectroscopic signal for registration with a point cloud over scanned surface.
    \item Iterative formation of 4D spatial-spectral point clouds.
\end{itemize}

\section{Related Works}
\label{sec:related_works}
\subsection{Spectral Quality Inspection}
Spectral sensing has enjoyed widespread use in many industries, but chiefly in the realm of precision agriculture. The presence of the \textit{red-edge} effect, caused by chlorophyll reflectance between 690-730 nm, has long been used to quantify the health of vegetation \cite{gupta2003comparative}. By analyzing various parts of the reflectance spectrum, researchers developed methods to detect plant ripeness \cite{shao2020assessment}, stress \cite{behmann2014detection}, disease \cite{rumpf2010early}, and phenotype \cite{liu2020hyperspectral}.

Classification of spectral signals via machine learning has been demonstrated in a variety of use cases, including pharmaceutical~\cite{luypaert2007near}, medical~\cite{bigio1997ultraviolet}, and recycling~\cite{eisenreich2006infrared}. The food industry relies on NIR spectroscopy to estimate the abundance of sugar, fat, and the overall quality of foodstuffs~\cite{kawasaki2008near, shah2020towards, guermazi2014investigation}. In these approaches, spectral signatures from a point spectrometer or imaging array are passed into a model that generates class probabilities representing the likelihood of the current signal belonging to each class. Another machine learning strategy attempts to compare unknown spectral signatures against a dictionary of reference spectra. These libraries have been extensively compiled with thousands of known materials and objects and are independent of the measuring spectrometer~\cite{kokaly2017usgs}.

\subsection{Spectral-Spatial Point Clouds}

Spectral information alone is not necessarily enough to adequately describe a complex object. HSI datacubes contain 2D spatial information, but constrain objects to a single plane of acquisition. Combining spectral information with point clouds allows a geometric interpretation of material information. \cite{debes2014hyperspectral} showed significant improvements in terrain classification by fusing remotely sensed hyperspectral datacubes and with LIDAR point clouds.
\cite{aasen2016acquisition, neuhaus2018high,honkavaara2012hyperspectral} fused snapshot hyperspectral data with LIDAR to create spatial-spectral digital surface models, but only with objects at a distance. \cite{edmonds2019generation,edmonds2020auto} also created hyperspectral point cloud data, but used an arm-mounted HSI to acquire hyperspectral images with optimizations for successive arm placements. To the best of our knowledge, no general approach has been developed to fuse high-resolution point spectral measurements with point clouds at close distances.
\subsection{Robot Spectroscopy}
The use of spectroscopy in robotics has gained traction as a way to classify household objects \cite{erickson2020multimodal}, survey geologic sites \cite{thompson2011autonomous}, explore underwater \cite{li2023underwater}, and assess fruit ripeness \cite{cortes2017integration}. Our previous work has been concerned with making pregrasp inferences using point spectral measurements \cite{hanson2022slurp, hanson2022vast, Hanson2022-Flexible}. However, in all these works we have assumed a fixed operating distance from object to sensor aperture, which simplified modeling of spectral signals. Further, the design and operation of these sensors assumed that a single-point scan was representative of the entire item composition, which is only true for select tasks in object manipulation. 
 
To analyze spectral signatures of complex objects, we require precise spatial data. A robotic platform can position a point spectrometer at known orientations to construct a 3D map. For example, Perseverance's X-Ray Fluorescence Spectrometer is positioned near a rock using a robotic arm, and a hexapod stage fine-tunes the sensor's pose \cite{allwood2015texture}. We seek to adapt this concept into an accessible platform for the general robotics community. 
\section{System Design}
\label{sec:system_design}
PROSPECT is a robot spectroscopy system that incorporates mechanical and optical elements for accurate formation and registration of spectral signatures with scanned geometries. PROSPECT uses a compact 6-degree-of-freedom (DoF) Stewart platform for precision orientation of a spectroscopy sensor suite relative to a measured surface. In this study, a 6-DoF robotic arm (UR3e, Universal Robotics) provides global motion around an object, while PROSPECT provides fine surface normalization and offsetting.
\subsection{Mechanical}
PROSPECT is designed to function as a precision end-effector adaptable to global motion platforms such as robot arms, gantries, or drones. PROSPECT contains a sensor suite with a fiber-optic cable, illuminating lamps, and time-of-flight sensors. These are mounted on the top platform of the Stewart mechanism, which allows for both spherical motion (for surface normal matching) and linear motion (for surface offsetting). This enables fine adjustment of the fiber-optic cable without the need to move the full weight of the global motion stage. The smoothness of spherical rotation also minimizes damage to the fiber optic cable that cannot be pinched or severely curved.


\begin{figure*}[!tbp]
    \centering
    \vspace{0.5em}
    \includegraphics[width=0.98\textwidth]{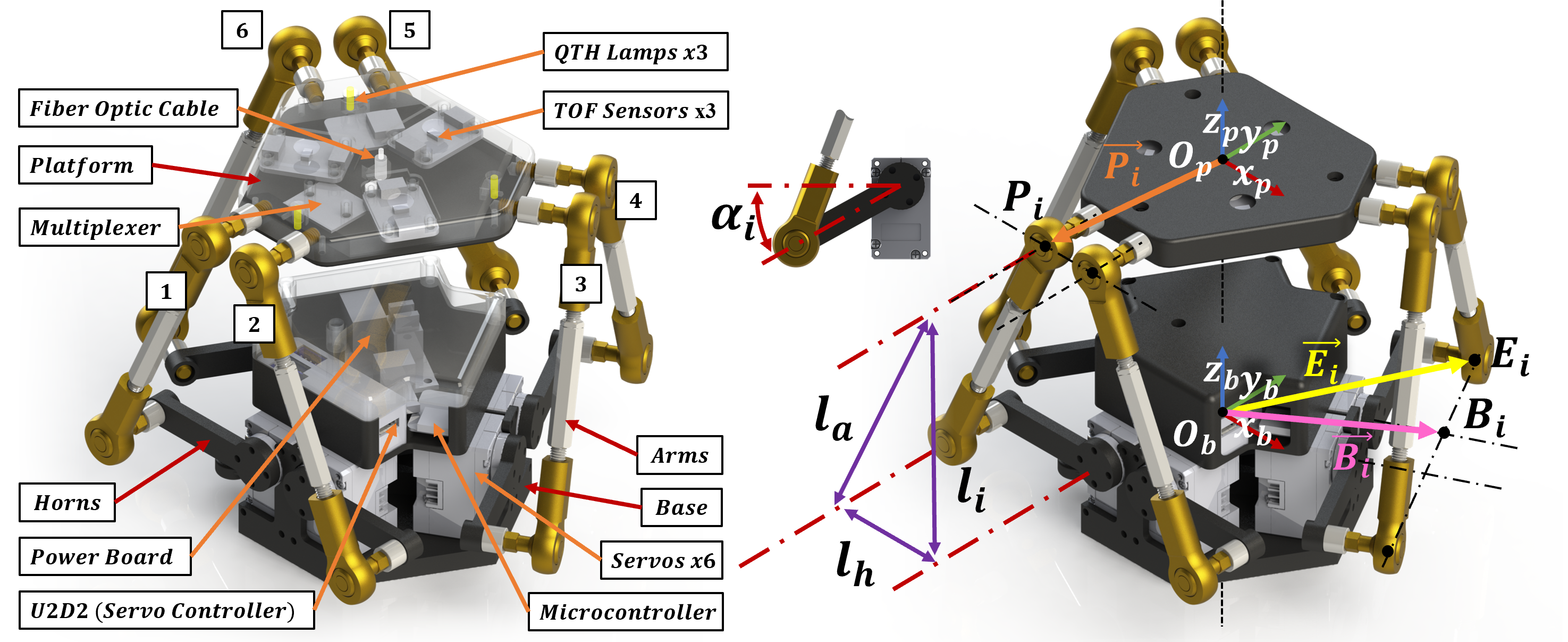}
    \caption{(Left) PROSPECT end-effector component diagram showing subcomponents. Individual motors and linkages are numbered. (Right) Identification of parameters and axes for kinematics and motion planning.}
    \label{fig:ortho}
    \vspace{-1.5em}
\end{figure*}

The PROSPECT end-effector is composed of a rigid base, a series of parallel links each composed of a horn $l_h$ and an arm $l_a$, leading to a sensorized platform as shown in Fig.~\ref{fig:ortho}. The base is 3D printed in Onyx on a Markforged X7 and contains wire routing and mounting to connect to a standard interface (Universal Robotics). Six servo motors (XL330-M288-T, Dynamixel) are mounted to the base. Each servo is oriented radially outward in a triangular pattern to maximize the motor packing density. The small angle between neighboring servos, $\theta_b$, is measured from the $z$ axis to the virtual point at the intersection between the servo axis and the plane of the arms. 

The base also holds a small 3D printed electronics enclosure with a microcontroller (QtPy, Adafruit), a servo controller, and servo power board. A custom servo horn leading to the anchor point of the ball-ended arm is attached to each motor. Each ball end arm consists of one left and one right threaded rod end ball joint joined by an aluminum turnbuckle to allow adjustment of the lengths of the arms for different desired workspaces. The upper ball joint of each arm connects to the outer edge of the top platform, as shown in Fig.~\ref{fig:ortho} below. The angle between the platform connection points, $\theta_p$, is measured from the z axis to the center of the upper ball joint. 
\subsection{Optical}
The top platform contains the sensors and electronics that enable point spectroscopy. Current spectrometers are too fragile and bulky to be fully encapsulated in the top platform; instead, the center of the platform contains a shielded fiber optic cable (ThorLabs) to guide light from the field of view of the platform and into a spectrometer at the base of the manipulator. The cable contains a 400 $\mu$m silica glass core with a numerical aperture (NA) of 0.50. The top platform clamps the ferrule end of the fiber to a fixed normal orientation relative to the platform surface. The fiber optic cable from the top plate connects to a VNIR diffraction grating spectrometer (Pebble VIS-NIR, Ibsen Photonics) with sensitivity from 500-1100 nm in 256 discrete wavelength samples. PROSPECT pairs with any grating spectrometer operating in the VIS-SWIR range (400-2200 nm); thus, it is reconfigurable for sensitivity to either a broad or narrow range of the electromagnetic spectrum given the dictates of the inspection application. 

Before operation, the spectrometer system is calibrated with a Spectralon reference standard and a dark noise collection according to \cite{lawrence2003calibration}.

\begin{equation}
\label{eq:spec_cal}
S_{cal spec,t} = \frac{S_{spec,t} - {\tilde{D}}_{spec,t}}{{\tilde{L}}_{spec,t} - {\tilde{D}}_{spec,t}}
\end{equation}
In (\ref{eq:spec_cal}), the tilde operator refers to the median of the spectral samples, collected over 10 measurements. The subscript $t$ indicates all measurements were collected with the same integration period. ${\tilde{L}}_{spec,t}$ refers to the median measurement taken from the Spectralon target at the nominal viewpoint distance. ${\tilde{D}}_{spec,t}$ refers to the median measurement taken with the spectrometer capped to incoming light. This reading accounts for dark current and other thermally induced electrical noise. ${S_{spec,t}}$ refers to the raw spectral signal for a particular sample and ${S_{calspec,t}}$ is the calibrated output.

The top platform contains 3 miniaturized Quartz Tungsten Halogen lamps (Ocean Insight). The lamps bathe the target surface in full-spectrum illumination, providing a sufficient signal for spectroscopic measurements. The lights operate at 5V and are controlled with a MOSFET circuit.

The top platform also contains 3 time-of-flight (ToF) distance sensors (STMicroelectronics VL53L4CD) oriented at 120 degree increments offset from the QTH lamps. The ToF sensors emit a pulse with a half-angle of 9$^\circ$. Given their spacing, the platform may use the three sensors to approximate a surface normal to the target surface at distances less than 6.0 cm. In this current work, these sensors are used to confirm the distance from the fiber ferrule tip to the object of interest.

Controlling the QTH lamp power is important to avoid interference with the ToF distance sensors, which measure distance from the surface with a 940 nm laser emitter. This wavelength is also emitted by the QTH lamps. For an optical acquisition cycle, the QTH lamps are switched off, the platform is oriented in the correct position using the ToF sensors for confirmation, the lamps are toggled on, and finally a spectral reflectance signal is integrated.

\section{Kinematics and Control}
\label{sec:kin_and_control}
\subsection{Inverse Kinematics}

The inverse kinematics of the Stewart Platform were derived to allow PROSPECT to achieve a given pose normal to a surface and a desired offset distance.

Given a pose of the Stewart platform described by a translation vector $\vec{T_{stew}} = \begin{bmatrix}x & y & z \end{bmatrix}^T$ and a rotation matrix $\mathbf{R_{stew}}$  describing a desired position relative to the base, there is a unique angle $\alpha_i$ to which servo $i$ must be rotated to reach the desired pose. $\vec{T_{stew}}$ is the displacement from the base origin $O_b$ to the platform origin $O_p$; $\mathbf{R_{stew}}$  represents the rotation of the platform with respect to the base. In this work, $\mathbf{R_{stew}}$ is defined in terms of Euler angles $(\psi, \theta, \phi)$ which represent an intrinsic rotation of $\phi$ about the x-axis, then $\theta$ about the y-axis, followed by $\psi$ about the z-axis.

\begin{equation}
\mathbf{R_{stew}} = \mathbf{R_z}(\psi)\mathbf{R_y}(\theta)\mathbf{R_x}(\phi)
\end{equation}



The vector $\vec{B_i}$ is the vector from the base origin $O_B$ to the point $B_i$. $\vec{P_i}$ is the vector from the platform origin $O_P$ to the point $P_i$. $\vec{q_i}$  is the vector from the base origin to the point $P_i$.

\begin{equation}
\vec{q_i} = \vec{T_{stew}} + \mathbf{R_{stew}} \cdot \overrightarrow{P_i} \label{eq:q_i}
\end{equation}

To solve the inverse kinematics for a Stewart platform actuated by rotational servomotors it is useful to solve the inverse kinematics of a Stewart platform actuated by linear servomotors. We consider $\vec{l_i}$, the leg which connects $B_i$ to $P_i$ in a linearly actuated Stewart platform.

\begin{equation}
\vec{l_i} = \vec{q_i} - \vec{B_i} \label{eq:l_i}
\end{equation}

We introduce one more parameter, $\beta_i$, which is the angle that the plane of rotation of servo $i$ makes with the x axis. Given this angular offset, it can be shown that the inverse kinematic equation is as follows using the derivation in~\cite{eisele_stewart_ik}.









\begin{equation}\label{eq:alpha_i}
    \alpha_i = sin^{-1}\Bigg(\frac{C_i}{\sqrt{A_i^2 + B_i^2}}\Bigg) - atan2(B_i, A_i) 
\end{equation}

where:

\begin{align}
    A_i &= 2l_hl_{i,z} \label{eq:A_i} \\
    B_i &= 2l_h(cos(\beta_i)l_{i,x} + sin(\beta_i)l_{i,y}) \label{eq:B_i} \\ 
    C_i &= l_i^2 - (l_a^2 - l_h^2) \label{eq:C_i}
\end{align}

To summarize, the inverse kinematic procedure -- solving for $\alpha_i$ -- is:
\begin{enumerate}
    \item Given a desired pose, determine $\vec{T_{stew}}$ and $\mathbf{R_{stew}}$ 
    \item Compute $\vec{q_i}$ from (\ref{eq:q_i}) and $\vec{l_i}$ from (\ref{eq:l_i})
    \item Solve (\ref{eq:A_i}-\ref{eq:C_i}) for $A_i$, $B_i$, and $C_i$
    \item Solve for $\alpha_i$ using (\ref{eq:alpha_i})
\end{enumerate}

\subsection{Validation of Inverse Kinematics}
\begin{figure}[b]
    \centering
    \includegraphics[width=\linewidth]{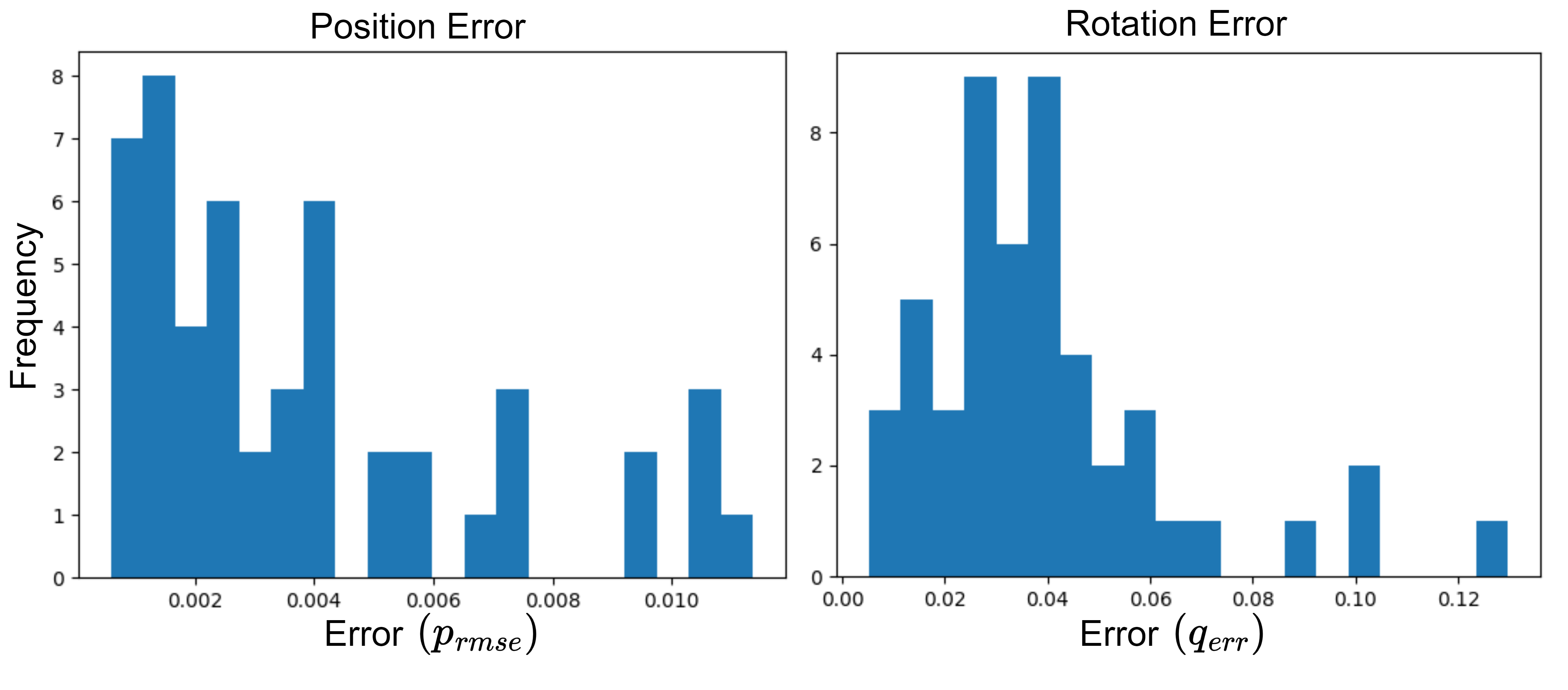}
    \caption{Histograms of errors from motion capture pose tracking.}
    \label{fig:kinmatic_validation}
\end{figure}
We validate the inverse kinematics of the PROSPECT platform using a motion capture camera (OptiTrack V120 Trio) to compare the commanded pose to the empirically observed pose. The platform was commanded to take uniformly sampled random poses for $z$ in the range of $\pm 20$ mm and $\pm 0.4$ radians in the $\theta,\phi$ axes from the neutral position.

To fully quantify the end-effector pose error, we report errors in position and orientation. Position error is given as the root mean squared error (RMSE) between the motion capture system's observed 3D coordinate $(x_1,y_1,z_1)_k$ and the commanded pose $(x_2,y_2,z_2)_k$ for sample $k$ out of $n$ total randomly sampled poses. The RMSE metric is given as:
\begin{equation*}
    p_{\mathrm{rmse}} = \sqrt{\sum_{k=1}^{n}\frac{(x_{1,k}-x_{2,k})^2+(y_{1,k}-y_{2,k})^2 + (z_{1,k}-z_{2,k})^2}{n}}.
\end{equation*}
3D rotation error is defined as the inner dot product of the ground truth and commanded unit quaternions, as defined in~\cite{huynh2009metrics}. Let OptiTrack's ground truth frame orientation be $\mathbf{q}_{\mathrm{1},k}$, and the corresponding commanded quaternion be $\mathbf{q}_{\mathrm{2},k}$. The error in rotation ($q_{\mathrm{err},k}$) for synchronized sample $k$ is defined over the range $[0,\pi]$ as:
\begin{equation*}
    q_{\mathrm{err},k} = \frac{\arccos{(|\mathbf{q}_{\mathrm{1,k}} \cdot \mathbf{q}_{\mathrm{2,k}}|)}}{n},
\end{equation*}
Fig.~\ref{fig:kinmatic_validation} shows the results of the kinematics validation experiments. Overall, the system is found to have a kinematic accuracy of $3.73 \pm 3.10$ mm in positional errors and $0.036 \pm 0.025$ radians in rotational accuracy. This error arises primarily from motor and bearing backlash. The backlash's effect on the platform varies based on its pose, but the positional and rotational error places an upper bound on the magnitude of the backlash of the system. The system's minimal angular error indicates precise surface normal matching is possible.

\begin{algorithm}[!b]
    \caption{View Point Planning}
    \textbf{Inputs}: (Point Cloud $([x_0, y_0, z_0]\hdots[x_n, y_n, z_n])$, \\
    Upwards Unit $(x, y, z)$ // $(0, 0, 1)$ \\
    Camera To Arm Base Transform, \\
    Voxel Size, // 1 cm \\
    Optimal Scanning Dist, // 3 cm \\
    Arm Offset // [0, 0, 5cm] 
    )
    
    \textbf{Begin}
    \begin{algorithmic}
    \label{alg:viewpoint_planning}
    \STATE{downsample\_cloud $\leftarrow$ \textit{downsample}(clean\_cloud, Voxel Size)}\\ 
    \STATE{normals $\leftarrow$ \textit{approximate normals}(downsample\_cloud)}\\
    \STATE{arm\_frames $\leftarrow$ [], stewart\_poses $\leftarrow$ []} \\
    \FOR{point $\in$ downsample\_cloud, normal $\in$ normals}
       \STATE{vector $\leftarrow$ \textit{normalized vector}(normal, point)} \\
       \STATE{viewpoint $\leftarrow$ Optimal Scanning Dist $\cdot$ vector}
        \STATE{arm\_point $\leftarrow$ viewpoint + Arm Offset}
        \STATE{stewart\_poses\textit{.append} ($\vec{T_{stew}}$ = viewpoint - arm\_point, $\mathbf{R_{stew,  z}} \parallel$ normal)}\\
       \STATE{arm\_frames\textit{.append}($\vec{T_{arm}}$ = arm\_point, \\$\mathbf{R_{arm,  z}} \parallel$ Upwards Unit)}\\
    \ENDFOR
    \STATE{\textit{transform}(frame $\forall$ frames $\in$ arm\_frames, \\Camera to Arm Base Transform)}
    \RETURN{arm\_frames, stewart\_poses, downsample\_cloud}
    \end{algorithmic}
   \label{viewpoint_planning_end}
\end{algorithm}
\section{Viewpoint Planning}
\label{sec:motion_planning}
With a point spectrometer, only a small portion of the object of interest is characterized per measurement. To characterize the entire object, we determine a subset of points that covers the surface of the object. We assume the existence of a point cloud that forms the basis for our viewpoint planning. To maximize scan consistency and spectral signal strength, the fiber optic cable is placed at a fixed distance from points of interest, as well as normal to the surface at that point.

\subsection{Point Cloud Preprocessing}
The raw point cloud from the 3D ToF Sensor (Azure Kinect, Microsoft) mounted above the arm contains points belonging the object of interest, as well as other extraneous points. 
Open3D \cite{open3d} is used for the point cloud processing. The extraneous points are removed through a series of preprocessing steps. First, the cloud is cropped based off of empirically determined global bounds. Then, the table upon which the object rests is discarded by removing inliers belonging to a plane found with RANSAC \cite{fischler1981random}. Finally, noisy points are removed, and points belonging to the largest object remaining in the scene are selected using Density Based Clustering \cite{ester1996density}. 

\subsection{Approach Positioning}

With the processed point cloud, viewpoints are selected using Algorithm~\ref{alg:viewpoint_planning}. Consistently spaced points that cover the viewable surface are found through voxel-based uniform down-sampling. The normal is then estimated at each downsampled point. For each point and corresponding normal, the Stewart platform is positioned at a fixed distance from the surface along the normal such that the fiber optic probe is pointed parallel to the normal, facing the point. The arm is positioned at a point with only a Z-axis offset to the Stewart platform, with a fixed rotation for all points, so that Algorithm~\ref{alg:viewpoint_planning} generalizes to the possible configuration of a Stewart platform affixed to the end of a 3-DoF system such as an $XYZ$ Gantry. Our experiments use the MoveIt motion planning backend to achieve desired approach positions \cite{coleman2014reducing}. 

\subsection{Surface Normal Matching}
To orient the fiber optic normal to the surface at a fixed depth, we implement a surface normalization algorithm. Given the initial pose of the Stewart platform, the positioning and readings of three ToF sensors onboard the end effector, and a target offset distance, a new pose for the platform is computed. The platform is then commanded to move to this position using the joint angles computed by the inverse kinematics algorithm.
\section{Spectral-Spatial Modeling}
\label{sec:spatial_spectral}
Despite their slender form, fiber optic cables acquire light from the environment in a more complex manner than tracing a ray from the fiber terminus to the surface. We begin by modeling the interaction of light reflected from a target object and into the ferrule of the fiber optic cable. Numerical aperture (NA) defines the maximum angle from which light will be accepted into the fiber and is given by: $NA = n_0\sin{\theta_{max}}$ where $n_0$ defines the refractive index of the medium through which light transits before entering the fiber, which we assume to be air with a refractive index of 1.0. $\theta_{max} = \arcsin{NA}$ yields the half-angle forming the acceptance cone. 

\begin{figure}[t]
    \centering
    \includegraphics[width=0.90\linewidth]{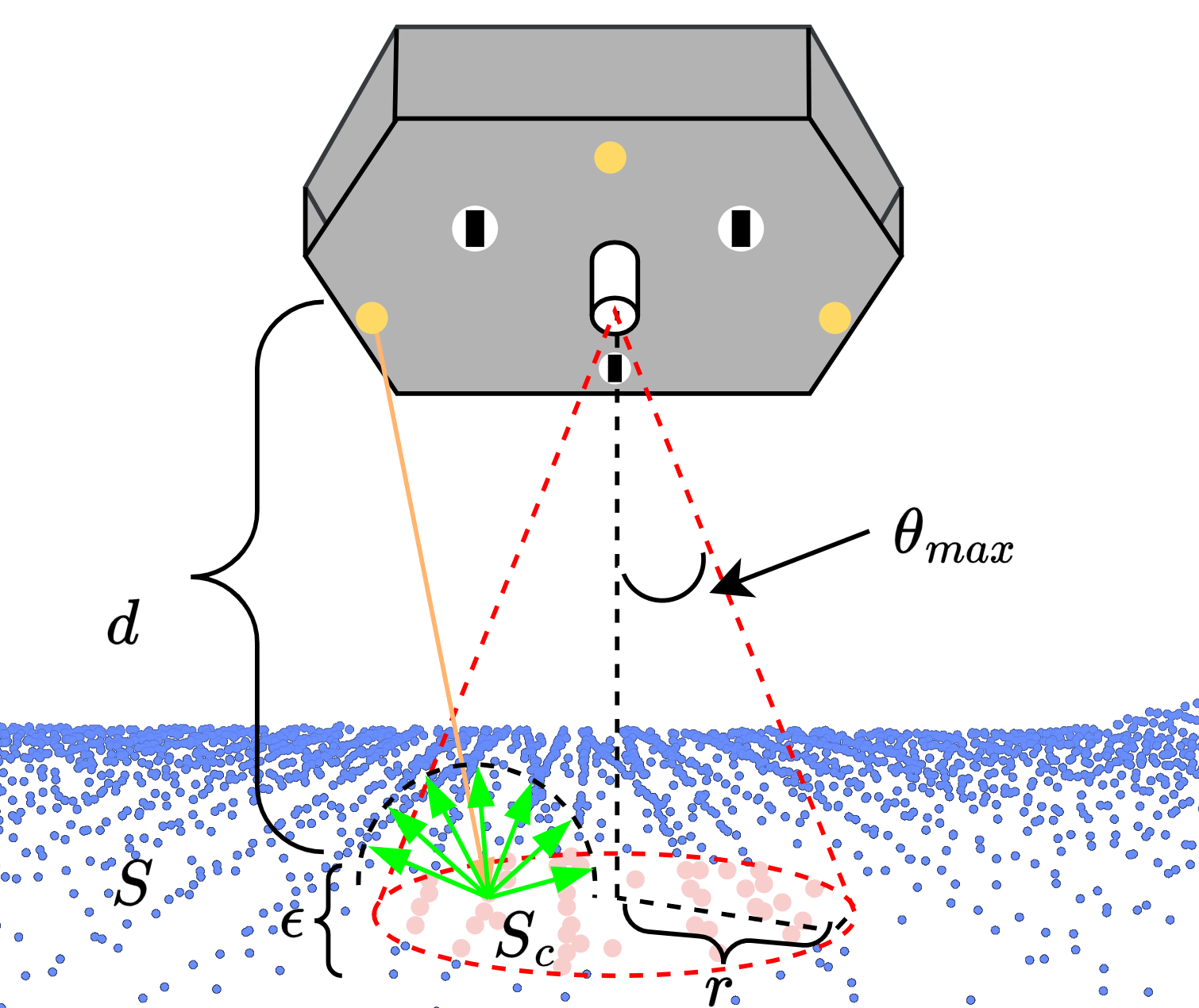}
    \caption{Model of spectroscopic measurement and modeling of object reflectance. Light is emitted from the PROSPECT platform (yellow) and reflected (green) by the surface $S$ (blue). The conical acceptance profile is defined by offset distance from the surface $d$, subsurface penetration $\epsilon$, and acceptance angle $\theta_{max}$. The intersection of the acceptance cone with the point cloud yields a subset of points $S_c$ (pink) with which the observed spectral signature is associated.}
    \label{fig:spectral_modeling}
    \vspace{-1.5em}
\end{figure}

Fig.~\ref{fig:spectral_modeling} shows the fiber optic cable above a nominal surface $S$ that the PROSPECT platform will scan. The acceptance cone is centered on the ferrule. The distance from the tip of the ferrule to the surface is given as $d$, which intersects with the surface in a circle with a radius $r$. $d$ is approximated from a static 3D scan of the object surface and confirmed using the 3 ToF distance sensors. The size of the cone base varies as a function of $d$, but $\theta_{max}$ remains constant. The area of the circular intersection region is given as $A = \pi (d \tan{(\arcsin{(NA)}}))^2$.

\begin{algorithm}[!tb]
    \caption{Spectral-Point Cloud Association}
    \textbf{Inputs}: (point\_cloud $([x_0, y_0, z_0]\hdots[x_n, y_n, z_n])$, \\
    Numerical Aperture $NA$,\\
    ToF Sensor Readings $[d_1, d_2, d_3]$,\\
    Subsurface Distance $\epsilon$,\\
    Radial Sampling $\rho$\\
    Stewart Pose $x_s,y_s,z_s$,\\
    Stewart Orientation $\psi, \theta, \phi$)
    
    \textbf{Begin}\\
    \begin{algorithmic}
     \label{alg:conic_intersection}
    \STATE{d $\leftarrow$ max($[d_1, d_2, d_3]$)}\\
    \STATE{cone\_points $\leftarrow$ \textit{create\_cone}(NA, d + $\epsilon$, $\rho$)} \\
    \STATE{rot\_cone $\leftarrow$ cone\_points$* R_{\psi} * R_{\theta} * R_{\phi} + [x_s,y_s,z_s]$
    \STATE{cone\_hull $\leftarrow$ \textit{convex\_hull}(rot\_cone)}}
    \STATE{triangulation $\leftarrow$ \textit{Delaunay}(cone\_hull)}
    \STATE{scanned\_points $\leftarrow$ []}
    \FOR{point $\in$ point\_cloud}
       \IF{\textit{contains}(triangulation, point)}
        \STATE{scanned\_points.\textit{append}(point)}
       \ENDIF
    \ENDFOR
    \RETURN{scanned\_points}
    \end{algorithmic}
\end{algorithm}

\begin{figure*}[!tbp]
    \centering
    \vspace{0.5em}
    \includegraphics[width=0.95\textwidth]{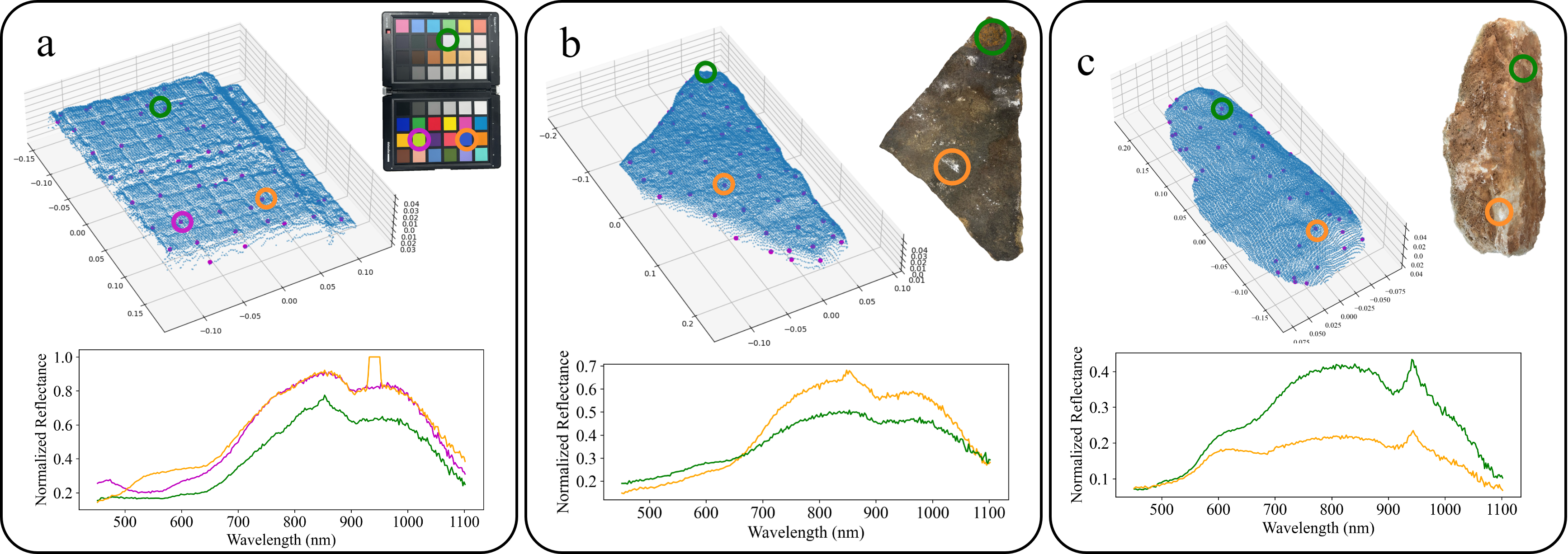}
    \caption{Experiments for sparse spectral-spatial modeling. Each subfigure also includes an RGB camera image of the scene, point cloud measurement, and spectral reflectance curves for extracted voxels. (a) Calibrated color checkerboard (Spyder) (b) Triangular sandstone slab (c) Gypsum boulder with Iron Oxide deposits.}
    \label{fig:experiments}
    \vspace{-1.5em}
\end{figure*}
To mark points as having been explored, we find the intersection between the surface $S$ and the circular profile of the acceptance cone projected on the surface, $S_c$. The intersection is found by creating a 3D mesh of the acceptance cone projected slightly below the surface of $S$, at a distance $\epsilon$. The mesh is then transformed into a Delaunay triangulation, which is queried to find points from $S$ that lie inside it, $S_c$. Since scanning occurs over the surface at a close distance, the number of points observed follows the following relationship: $ 1 \leq ||S_{c}|| << || S || $. As the end-effector is translated according to Algorithm~\ref{alg:viewpoint_planning}, the point cloud is annotated so that each point $x_i,y_i,z_i$ in $S_c$ is associated with a spectral vector $[\lambda_0,\lambda_1,\dots,\lambda_n]$.

As the size of $A$ increases, the intersecting area and points in $S_c$ also increase. However, as the distance $d$ increases, the number of points contributing to the spectral signal similarly increases, while the overall intensity of the signal decreases because fewer photons enter the fiber probe. This mixing of spectral signatures from multiple points complicates the interpretation and classification of the signal. Knowing the approximate point density of the point cloud $S$, we chose a constant value for $d$ to avoid variable signal quality. Algorithm~\ref{alg:conic_intersection} details the steps to extract the points $S_c$ and form the correlated spectral-spatial measurements.

\section{Experiments}
\label{sec:experiments}
We validate the results of PROSPECT by acquiring spectral-spatial point clouds of objects with complex color and material composition, and highlight the need for accurate viewpoint planning and precise control of the fiber optic probe.

\subsection{Sparse Point Acquisition}
The first set of experiments demonstrates the creation of a sparse spectral-spatial point cloud. These scans are designed to be completed in less than 5 minutes to broadly understand an object's surface composition. Fig.~\ref{fig:experiments} shows the variation of spectral signatures across the surface of an object and emphasizes the importance of collecting precise spectral measurements. In each of these scenarios, viewpoints were planned with a voxel size of 5 cm. 

We test our viewpoint planning on a calibrated color checkerboard, a triangular sandstone slab, and a Gypsum boulder with Iron Oxide deposits as seen in Fig.~\ref{fig:experiments}. The color checker is a highly diverse surface of tiled color squares with varying albedo. The rocks used in this experiment are notional examples of complex geometries derived from inspection and identification of target minerals. For each viewpoint, if planning is not feasible for either the arm or Stewart platform due to potential collisions or unreachable poses beyond the operable workspace, we discard that viewpoint and continue.  
\subsection{Point Alignment}
From the results in Section~\ref{sec:kin_and_control}, we know PROSPECT has minimal pose error. Building on these results, we design two experiments to explore the precision of the device. Using the color checker from the previous experiment, we plan a trajectory to directly over a red-colored square at the nominal working distance. After acquiring a sample, the arm is linearly translated with PROSPECT in the same orientation along the upward $z$ axis by 12 mm. The arm is then perturbed by adding a small angular offset to the rotation in two increments of 11 and 17 degrees. This set of experiments is designed to highlight small errors induced by sample-based motion planning methods.

Additionally, we explore the necessity of the Stewart platform to construct accurate surface profiles by comparing the spectral signatures acquired from the surface normal using PROSPECT compared to those acquired from the target offset distance, but along the positive $z$ axis. Due to the complex geometry of the rock, we avoid planning solely with the 6-DoF manipulator as many generated arm configurations are infeasible or would damage the optical equipment.
\section{Results \& Discussion}
\label{sec:results}

\begin{figure}[!tbp]
    \centering
    \includegraphics[width=\linewidth]{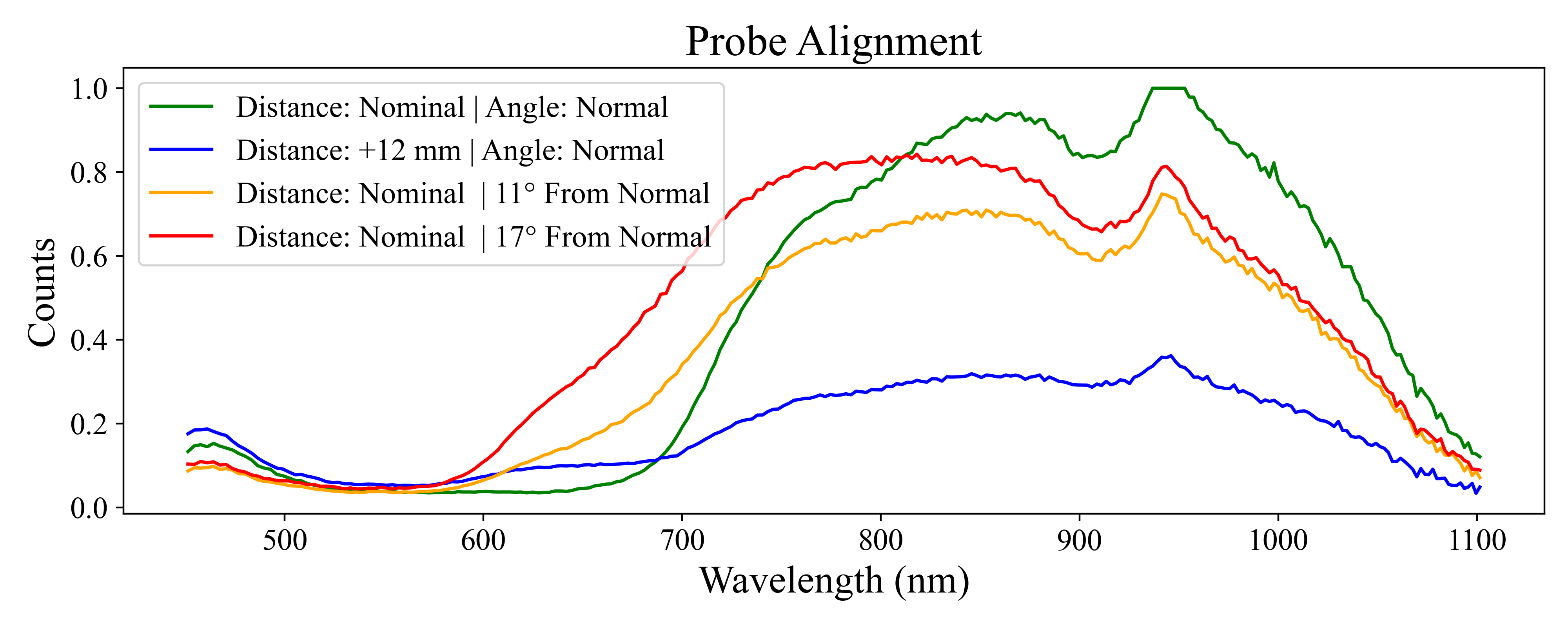}
    \caption{Reflectance profiles for different probe alignments with fixed $T_{stew, xy}$.}
    \label{fig:alignment_verification}
    \vspace{-1.0em}
\end{figure}

The results of the probe alignment underscore the need for precision alignment with fiber optics.  Fig.~\ref{fig:alignment_verification} shows the reflectance profiles collected in each of the scenarios. The nominal reflectance profile shows a markedly different shape from each of the other test scenarios. In previous research \cite{hanson2022slurp}, we assumed a uniformity in the composition of the scanned surface and expected the magnitude of the spectral signal to vary linearly as a function of distance from the surface. As modeled through Algorithm~\ref{alg:conic_intersection}, as the distance increases, the number of points contributing to the spectral signature also increases, but at a quadratic rate. The increased scan height results in adjacent color cells contributing to the observed reflectance profile. Similarly, when the probe is rotated out of alignment by small angles, the conic intersection occurs at an oblique angle, with a larger spread of points beyond the intended intersection area.

\begin{figure}[!tbp]
    \centering
    \includegraphics[width=\linewidth]{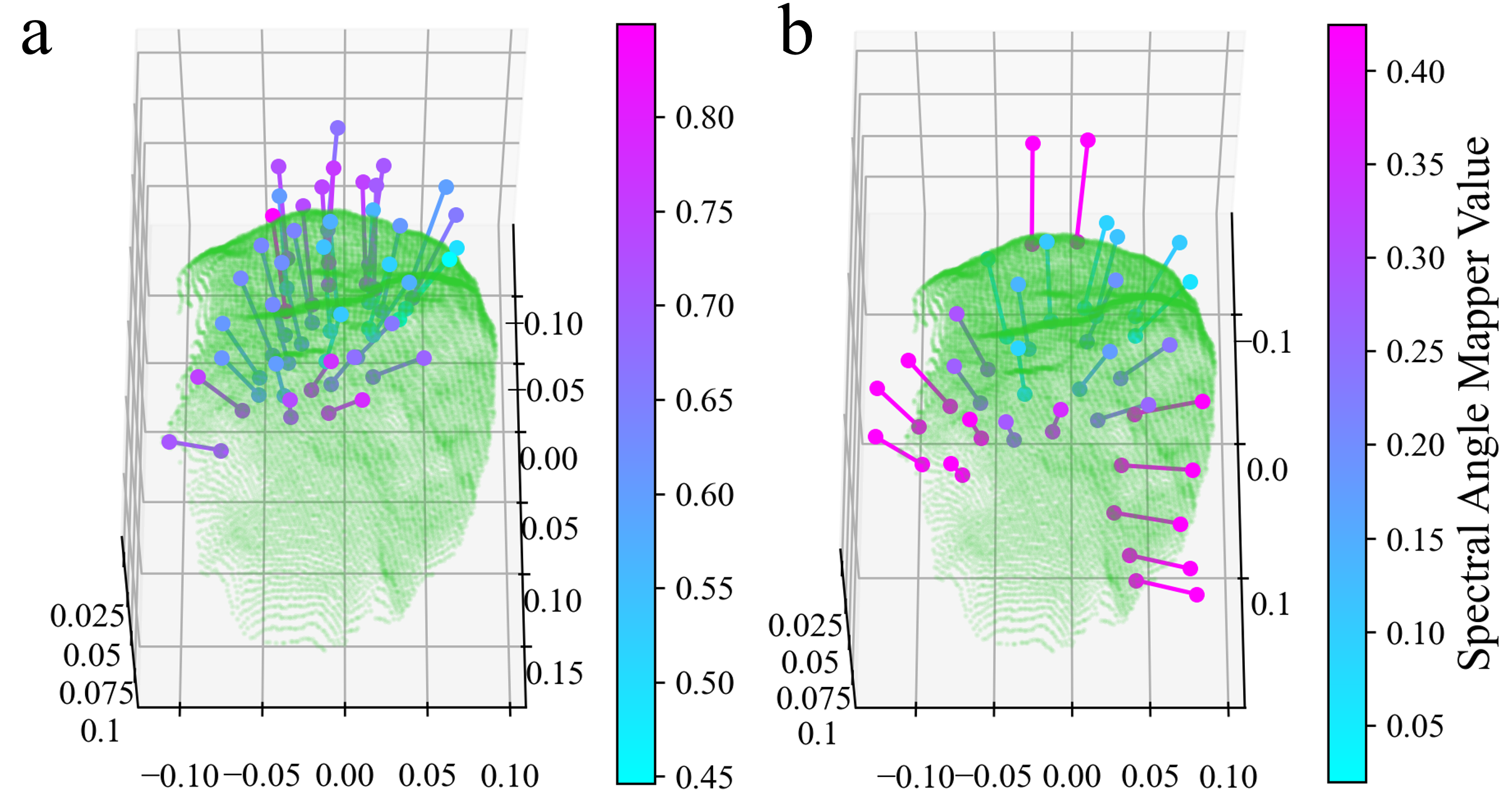}
    \caption{Point cloud of gypsum boulder with sampled viewpoints and corresponding normals, colorized by their SAM score compared to PROSPECT using the method proposed in Algorithm~\ref{alg:viewpoint_planning}. (a) Compared to viewing points directly from above with no rotation. (b) Compared to where Stewart platform always points straight down, which may not be parallel to the surface normal.}
    \label{fig:prospect_ablation}
    \vspace{-1.75em}
\end{figure}
Both sources of imprecision lead to scenarios where the observed points may constitute a mixture of multiple spectral signatures. Further research into spectral mixing of observed measurements will help to understand if a set of scanning parameters will induce imprecision.

 While Fig.~\ref{fig:alignment_verification} demonstrates the errors induced by small misalignments in motion planning, we also quantitatively justify the use of PROSPECT to match surface normals over a complex geometry. We select two alternative approaches to demonstrate the effectiveness of our end-effector: one where all normal vectors in Algorithm~\ref{alg:viewpoint_planning} are overwritten to be parallel to the unit vector, and one where in Algorithm~\ref{alg:viewpoint_planning} all $\mathbf{R_{stew}}$ are replaced with unit rotation. The first alternative approach is designed to emulate viewpoint planning for PROSPECT on a 3-axis gantry, and the second alternative approach is designed to emulate poor alignment of the spectrometer relative to an object's surface normal. As demonstrated in Fig.~\ref{fig:alignment_verification}, orienting the spectrometer to be parallel to the normal of the surface yields the strongest signal amplitude and as such is assumed to be optimal for downstream processing. The alternative approaches' spectral signatures differ from that of the normal matching baseline, suggesting that spectral profiling quality suffers without PROSPECT. We calculate the dissimilarity between any two collected point clouds using the Spectral Angle Mapper (SAM), with results shown in Fig.~\ref{fig:prospect_ablation}.
\begin{equation}
      SAM(s_p,\hat{s}) = \arccos{\left( \frac{
     s_p\cdot \hat{s}
    }{
    ||s_p|| \times || \hat{s} ||
    }\right)}
\end{equation}

SAM is well supported in spectroscopy literature as a means to calculate the divergence between two spectral signals as a similarity metric \cite{kruse1993spectral} while showing robustness to illumination intensity. In this formulation $s_p$ is the spectral vector observed using PROSPECT, and $\hat{s}$ represents the correlated spectra vector collected through one of the ablated motion planning routines. SAM is defined over $[0, \pi]$, where $0$ indicates a high degree of similarity and $\pi$ a high dissimilarity.

Fig.~\ref{fig:prospect_ablation} shows the dissimilarity in the acquired measurements. Fig.~\ref{fig:prospect_ablation}a presents the results of alignment along the unit vector compared to PROSPECT. Table~\ref{tab:sparse_results} highlights the mean, standard deviation, and range of the observed values. Both strategies exhibit a wide range of SAM scores, indicating that spectral measurements do not diverge uniformly across the surface. In both subfigures of Fig.~\ref{fig:prospect_ablation}, there is a corresponding decrease in similarity as the normal diverges from alignment with the $z$ axis. Points on a relatively flat surface exhibit close agreement with measurements scanned without spherical alignment since the platform was already neutrally aligned in those scenarios.

\begin{table}[t]
\vspace{1.0em}
\caption{Summary Statistics for Spectral Signature Similarity Compared to Baseline Viewpoint Planning}
\label{tab:sparse_results}
\centering
\footnotesize
\setlength\tabcolsep{4 pt} 
\begin{tabular}{c c c c}
\toprule
    \textbf{Approach} & \textbf{$\overline{SAM}$} & \textbf{$SAM_{\sigma}$} & range $SAM$\\
    \midrule
    Exclusively 3-axis Viewpoint Planning  & 0.651 & 0.088 & 0.403\\
    Poor Alignment to Object Surface Normal & 0.350 & 0.226 & 0.783\\
    \bottomrule
    \end{tabular}
    \vspace{-1.75em}
\end{table}


\section{Conclusions}
\label{sec:conclusions}
In this work, we presented PROSPECT, an integrated end-effector for the precise control of a fiber optic cable for automated spectral measurements of complex geometries. Our proposed motion planning and physics-based point attribution allows for the progressive construction of spectral-spatial point clouds. We demonstrated the validity of our approach in increasingly complex scenarios with variable spectral profiles. For generality to the robotics community, the system is designed to be agnostic to the robot and spectrometer model.

In future works, we plan to expand our approach of scanning objects to identify and segment non-visible material imperfections through the use of automated anomaly detection, as in our previous work \cite{hanson2022occluded}. New ToF sensors offer multizone ranging; allowing for the three ToF sensors to be condensed to a single unit and streamline the electronics in the top platform. Additionally, we will increase the speed of motion planning through coupled motion planning of the end-effector and PROSPECT for smooth continuous trajectories over object contours. We believe this work represents a practical step towards the fusion of spectral sciences and robotics and opens new opportunities to non-destructive characterization in manufacturing and automation.


\bibliographystyle{IEEEtran} 
\bibliography{references}





\end{document}